# Data-Optimized Contingency Screening: A Machine Learning Approach to Power System Security

Joshua Salako
*dept. of Electrical and Electronics Engineering*
*Federal University of Agriculture*
Abeokuta, Nigeria
0009-0005-1044-2616

Folajimi Osikomaiya
*dept. of Electrical and Electronics Engineering*
*Federal University of Agriculture*
Abeokuta, Nigeria
0009-0000-2998-3256

Olakorede Olamiju
*dept. of Electrical and Electronics Engineering*
*Federal University of Agriculture*
Abeokuta, Nigeria
0009-0009-1913-8265

***Abstract*—Ensuring the security of the power system is essential for stability and reliability, especially in the event of disruption. Effective classification of contingency in power systems enables proactive decision-making and mitigates large-scale breakdowns and failures. This study explores the use of machine learning algorithms to classify security levels of contingencies in power systems into safe, moderate or severe classes. For this approach, Newton-Raphson load flow method extracts system data from contingency scenarios, using Overall Performance Index (OPI) as safety measure. For data pre-processing, Synthetic Minority Over-Sampling Technique (SMOTE) and Principal Component Analysis (PCA) is used to address class imbalance and reduce dimensionality, respectively. K-Nearest Neighbours (KNN), Random Forest (RF) and Support Vector Machines (SVM) is trained and evaluated on datasets generated through N-k contingency scenarios for k equal 1, 2, and 3 on IEEE-14 and IEEE-30 bus systems using four hybrid pre-processing configurations: normalized, SMOTE-balanced, PCA-transformed, and a combined SMOTE PCA-transformed. Performance is assessed by precision, recall and F1 score, with priority given to the severe contingency classes. The RF achieved the highest F1 scores of 0.97 in IEEE-30 and 0.86 in IEEE-14, SVM benefits significantly from PCA and improves the accuracy of the classification, while KNN is best suited for SMOTE and PCA conversion. The findings show that PCA contributes more than SMOTE to the overall performance of the model. However, SMOTE improves recall but can introduce false positives and is therefore a compromise of accuracy. This study highlights machine learning as a scalable and powerful alternative to traditional contingency analysis, which improves the assessment of security in real time.***

## I. INTRODUCTION

A stable and reliable supply of electricity is of the utmost importance in modern society, as it plays a crucial role in facilitating the activities of the various sectors, including residential, commercial, industrial and infrastructure [1]. The primary objective of the energy system is to maintain a stable and adequate supply of electricity to customers, while respecting the fixed limits of frequency and voltage. Power systems, which consist of interconnected generation, transmission and distribution networks, ensure uninterrupted supply while respecting the limits of frequency and voltage [2]. However, the challenges are posed by energy system contingencies which pose significant obstacles to the smooth and reliable functioning of these systems [3]. The safety of the energy system is paramount, referring to the ability of the system to maintain stability and functionality during credible risks. The contingencies may be caused by a variety of factors such as equipment failure, unexpected faults, natural disasters, human error or deliberate attacks on infrastructure. Therefore, there is an urgent need for a robust detection process to determine whether a system is operating safely or whether it is in an emergency situation at any given time [4][20].

Contingencies have the potential to disrupt the normal operation of the energy system, leading to blackouts, voltage instability, cascading failures and consequent social and economic consequences [29]. The impact extends to sectors such as health, transport, communication, finance and general public security and welfare. Therefore, accurate identification, classification and understanding of energy system contingencies and their immediate response is essential for efficient contingency management, timely decision making and implementation of measures to restore system stability and minimise the impact on end-users [5]. Traditional energy system contingency classification is highly dependent on manual analysis and expertise [6]. This manual approach, carried out by experienced energy system experts, entails challenges such as time-consuming procedures, especially in large systems with many interconnected components and growing data. The increasing complexity of modern energy systems makes it difficult for professionals to analyse and classify in real time the contingencies in a comprehensive way, and introduces the risk of inconsistencies and bias. This underlines the need for automated and intelligent methods to help classify emergencies and provide accurate and timely assessments for efficient decision-making and response strategies.

Machine learning has emerged as a powerful tool in modern engineering and decision-making processes, offering data-driven solutions to complex problems in a wide range of areas[21]. Machine learning algorithms allow automatic pattern recognition, classification and prediction by analysing large datasets. These techniques have shown significant improvements in accuracy and efficiency in areas ranging from health care to finance and industrial automation [16]. One of the key applications of machine learning is in the security of power systems, where ensuring the stability and reliability of the electricity supply is of particular importance [29]. Power systems operate under dynamic conditions, which make them vulnerable to disturbances such as breakdowns of transmission lines, power failures and cascading failures. In order to mitigate these risks, contingency analysis is carried out to assess the resilience of the system in the event of a disruption. Traditional methods, such as deterministic safety assessments, are computationally intensive and often struggle to provide insights in real time [17].

The K-nearest neighbors (KNN) algorithm, a non-parametric machine learning technique, has garnered significant attention for its capability to classify contingency scenarios based on historical system data [30]. By analyzing patterns from past contingency cases, KNN can effectively predict event severity, enabling grid operators to implement preventive measures. Research indicates that machine learning methods substantially improve contingency screening and classification accuracy while reducing computational complexity without compromising reliability [18][19]. In addition to KNN, Random Forests (RF) and Support Vector Machines (SVM) have also proven effective in contingency classification for power systems. RF, an ensemble learning approach, constructs multiple decision trees during training and determines the class prediction through majority voting [31]. This methodology enhances classification robustness by mitigating overfitting and improving generalization, making it particularly suitable for high-dimensional power system datasets. RF has been widely employed in power system security assessment due to its ability to manage intricate variable interactions while delivering interpretable results for contingency ranking and classification.

Conversely, SVM, a powerful supervised learning algorithm, seeks to identify an optimal hyperplane that maximizes the margin between different classes in a high-dimensional space [32]. Particularly effective in both binary and multi-class classification tasks, SVM is a strong candidate for contingency classification. By leveraging kernel functions, SVM can transform non-linearly separable data into a higher-dimensional space, thereby enhancing classification performance in complex contingency scenarios [33]. The integration of machine learning-driven classification models enables power system operators to strengthen decision-making, optimize grid stability, and mitigate the risks associated with unforeseen contingencies. The increasing adoption of these approaches represents a paradigm shift in power system security, fostering more adaptive and intelligent decision-making frameworks [22].

Three machine learning algorithms—RF, SVM, and KNN—are investigated and assessed in this study for power system contingency classification in an effort to overcome the drawbacks of manual methods. The remainder of the work is organized as follows: Section II covers related works, and Section III displays the methodology, modeling, and data generation and preprocessing of the machine learning classification algorithms. Section IV concludes with a thorough discussion of the outcomes and conclusions drawn from the assessment of the research methodology.

## II. RELATED WORKS

The analysis of the contingencies, which is an essential activity in the planning and operation of the power system, involves three main stages: defining, selecting, and evaluating contingencies. The selection of a contingency, which includes ranking and screening methods based on an overall performance index, is traditionally a computationally complex task for real-time applications. The effective solution is to integrate traditional approaches using security indices with machine learning algorithms. This combined approach provides a more precise and practicable solution for real-time applications in the context of power system contingency analysis [8]

In Reference [9], the authors proposed an approach using online extreme sequence learning machines to identify power quality events in real time, significantly increasing the reliability of the power system. This method improves the monitoring and classification of quality events, which allows operators to identify and deal with interruptions quickly, thus ensuring a continuous supply of high quality electricity. A study [10] examined the integration of a Unified Power Flow Controller with the Extreme Learning Machine technology to improve the safety of transmission lines. This approach optimises real-time monitoring of power flows, alleviates congestion of transmission lines and improves the security and reliability of power transmission systems. In Reference [11], machine learning techniques have been used to assess the quality of electricity in the distribution networks. The method proposed improves the accuracy of the analysis of the quality of the electricity, allowing for rapid detection and mitigation of problems such as voltage variations and harmonics, thus increasing the reliability of the power distribution system. Reference [12] focused on improving the monitoring of the quality of power from microgrids with integrated photovoltaic power sources. By using machine learning techniques to identify and classify energy quality events accurately, this approach addresses the problems of intermittent solar energy and ensures that microgrids are stable and reliable. In Reference [13], a method based on deep learning was proposed to accurately estimate voltage drops in power systems with limited monitoring capability. This technique increases the speed and accuracy of voltage sag detection and provides a practical solution for maintaining the reliability of the electrical system within the monitoring limits. The active learning solution introduced in [14] has increased the use of machine learning in power systems by reducing the computational complexity of traditional methods. This approach gives priority to critical data for security assessment and improves the efficiency and accuracy of real-time monitoring and decision making in the operation of power systems.

### *A. Contingency Analysis of Power Systems*

Contingency analysis is a critical process in power system operations to enable grid operators to assess the impact of potential failures and take proactive measures to maintain system stability and security [36]. The process involves three main stages: contingency definition, selection, and evaluation.

*1) Definition phase:* In this phase, potential contingencies, such as line outages and equipment failures, are identified and characterized in terms of scope and severity.

*2) Contingency selection:* The purpose of the contingency selection is to select a subset of contingency for further analysis. This selection is usually based on criteria that prioritise the contingencies in terms of their potential impact on the security of the system. During this process, factors such as voltage or frequency disturbances, power flow variations and the criticality of the components concerned are taken into account.

*3) Evaluation stage:* During this phase, the selected contingency is thoroughly assessed in order to understand their implications for the security of the system. For the evaluation of steady-state and transient conditions, either power flow analysis or dynamic simulation is performed. Performance metrics, such as voltage fluctuations, power flow losses, and stability margins, will be computed to measure the intensity of every occurrence.

Contingency analysis provides insight into the vulnerability of the power system. To keep the system stable and secure, operators can take corrective and preventive actions, such as load shedding or control measures [15].

### B. Load Flow Analysis

Analysis of the load flow plays a key role in assessing the performance of the power system, optimising the operation of the network and identifying potential vulnerabilities in the voltage range, phase angles, active and reactive power of the different bus under steady-state conditions [37]. The load flow equation is a mathematical representation of the flow of power in a power system network. Various formulations may be used to represent network equations in power system analysis, the common approach is the node voltage method. When expressed in the form of nodal admittance, these equations result in a series of complex linear simultaneous algebraic equations involving node currents.

An electricity system of realistic size can consist of hundreds of buses and generators and thousands of lines. Newton-Raphson (NR) is an iterative technique with quadratic convergence, which makes it a mathematically better choice than the Gauss-Seidel method, especially for non-conditional problems. It has proven to be suitable for large energy systems, which require only a few iterations to reach a solution, regardless of the size of the system. However, this requires an evaluation of all functions in each iteration. Although most power flow problems converge in less than ten iterations, each iteration requires additional time and computer power [38]. The current injection at bus $i$ is expressed in (1).

$$I_i = \sum_{j=1}^{n} Y_{ij} V_j \tag{1}$$

where $I_i$ is the bus current $i$, $Y_{ij}$ is the admittance between buses $i$ and $j$, $V_j$ represents bus voltage $j$, and $n$ represents the total number of buses in the power system. By expressing (1) in polar form,

$$I_i = \sum_{j=1}^{n} |Y_{ij}||V_j| \angle(\theta_{ij} + \delta_j) \tag{2}$$

Equation (3) shows that the complex power at bus $i$ is to be:

$$P_i - jQ_i = V_i I_i \tag{3}$$

where $P_i$ is the active power at bus $i$, $Q_i$ is the reactive power at bus $i$. Substituting $I_i$ from (2) into (3) and separating the real and imaginary parts to obtain (4) and (5).

$$P_i = \sum_{j=1}^{n} |V_i||V_j||Y_{ij}| \cos(\theta_{ij} - \delta_i + \delta_j) \tag{4}$$

$$Q_i = -\sum_{j=1}^{n} |V_i||V_j||Y_{ij}| \sin(\theta_{ij} - \delta_i + \delta_j) \tag{5}$$

Equations (4) and (5) represent nonlinear algebraic relations in terms of independent variables, which are expressed in per unit (p.u.) with angles measured in radians. For every load bus (PQ), both equations apply, whereas for each voltage regulated bus (PV), only (4) is relevant.

By performing a Taylor series expansion of these equations around estimated starting conditions, taking slack bus as reference and disregarding higher-order terms, the outcome is:

$$\begin{bmatrix} \Delta P \\ \Delta Q \end{bmatrix} = \begin{bmatrix} \frac{\partial P}{\partial \theta} & \frac{\partial P}{\partial V} \\ \frac{\partial Q}{\partial \theta} & \frac{\partial Q}{\partial V} \end{bmatrix} \begin{bmatrix} \Delta \theta \\ \Delta V \end{bmatrix} \tag{6}$$

The load flow equations are solved by an iterative process in which the voltage magnitudes and phase angles of each bus are repeatedly updated until the specified convergence criterion has been reached. This process is repeated until the power balance equations are met and the system is in steady state operation. If convergence has not been reached then the voltages at the busbars are updated using:

$$\delta_i^{(k+1)} = \delta_i^{(k)} + \Delta \delta_i^{(k)} \tag{7}$$

$$|V_i^{(k+1)}| = |V_i^{(k)}| + \Delta |V_i^{(k)}| \tag{8}$$

where $\delta_i^{(k+1)}$ is the angle result for bus $i$ for iteration $k+1$, and $|V_i^{(k+1)}|$ is the voltage magnitude for bus $i$ for iteration $k+1$.

Emergency analysis using AC power flow has the advantage of providing information on power flow in terms of megawatts (MW), megavolt-amperes (MVAR) and bus voltage. This approach allows for the assessment of voltage-limit overloads and of precise voltage-limit failures. In this study, the focus is on power outages for ranking the contingencies and the performance indicators (PI) are taken into account to assess the severity of each contingency. Conventional methods of power flow are used for the calculation of these indices in the offline mode.

The obtained values through the conventional methods are then arranged in descending order, with the contingency having the highest PI value being ranked first. Two primary types of performance indices are typically employed for contingency analysis:

*1) Active Power Performance Index:* The active performance index denoted by $PI_P$ is essential for assessing the extent of line congestion. It identifies scenarios in which the active power on some transmission lines is above the limit value. It provides a quantitative assessment of the potential for congestion on the line. It does so by checking the flow of active power through the transmission lines. When active capacity on a given line approaches or exceeds its capacity, this indicates an increased risk of line congestion. This index is a key element in contingency analysis, where it helps to identify the contingencies that may result in an excessive active power flow. These contingencies, if not addressed, could compromise the operational stability of the energy system. The value assessment guides power system operators in the implementation of preventive measures. By identifying lines of higher severity, operators can take proactive remedial measures such as redirecting power flows, shedding loads or adjusting system parameters to mitigate the risk of overload. The active power performance index is expressed in (9).

$$PI_P = \sum_{i=1}^{N_L} \left(\frac{W}{2n}\right) \left(\frac{P_i}{P_{imax}}\right)^{2n} \tag{9}$$

where $P_i$ is the MW power flow of line i, $P_{imax}$ is the MW capacity of line $i$, $N_L$ is the number of lines in the system, $W$ is the real non-negative weighting factor (default value 1), and $n$ is the exponent of the penalty function (default value 1).

*2) Voltage performance index:* The voltage performance index denoted by $PI_V$ expressed in (10), assists in identifying violations of bus voltage limits. It enables evaluation of voltage levels at different buses within the system and provides insights into the voltage stability of the system.

$$PI_V = \sum_{i=1}^{N_B} \left(\frac{W}{2n}\right) \left(\frac{|V_i| - |V_{isp}|}{\Delta V_{ilim}}\right)^{2n} \tag{10}$$

where $|V_i|$ is the voltage magnitude at the $i^{th}$ bus, $|V_{isp}|$ is the rated voltage magnitude at the bus, $\Delta V_{ilim}$ is the deviation limit of the voltage, $n$ is the exponent of the penalty function (default value 1), $N_B$ is the number of buses in the system considered, and $W$ is the real non-negative weighting factor (default value 1).

Bus voltages are additionally responsive to the generation of reactive power by the generating units. $PI_V$ offers valuable insights into the seriousness of abnormal voltage levels, provided that the production of reactive power remains within permissible limits. In the event of a contingency, if the reactive power approaches its maximum limits, standard AC load flow analysis is employed to determine bus voltages. This process leads to deviations from the scheduled voltage at generator buses. Consequently, voltage analysis during contingencies necessitates consideration of the reactive power constraints imposed on the generators.

In power systems, voltage performance index is used to identify buses with notable voltage deviations, which may lead to voltage stability concerns or voltage collapse. Ensuring $PI_V$ values remain within acceptable limits will make the power system maintain stable and reliable voltage levels, thus ensuring smooth operation and minimizing the risk of voltage-related disturbances [35].

*3) Overall performance index:* Overall Performance Index (OPI) is a significant parameter used in power system contingency analysis to evaluate the system's security status when facing various contingency scenarios. It considers both Active Power Flow Performance Index and Voltage Performance Index, providing a comprehensive assessment of the system's capability to handle reactive power flows and maintain voltage stability.

OPI is derived by the addition of the Active Power Performance Index and the Voltage Performance Index shown in (11).

$$\mathrm{OPI} = \mathrm{PI}_P + \mathrm{PI}_V \tag{11}$$

OPI helps determine the security status of power systems. A higher OPI indicates a higher degree of uncertainty in the system. Therefore, when prioritising the contingencies, the focus is on those with higher OPI values. To facilitate a detailed analysis, the Newton-Raphson Load Flow (NRLF) solution is used to extract the system parameters for each N-1 line outage contingencies. The resulting OPI provides valuable insight into the security level of the system in a specific contingency scenario. A higher OPI indicates better security and robustness, while a lower value may indicate possible vulnerabilities or critical conditions that require immediate attention. By calculating OPI for different contingency scenarios, power system operators can prioritise their actions, identify critical contingencies and ensure the reliability and security of the power system under different operational conditions.

The flowchart of the load flow algorithm shown in Fig. 1 outlines the contingency screening process to evaluate power system security in this work. The algorithm begins with initializing and reading system variables, followed by performing a load flow analysis for the pre-contingency case. Once the baseline conditions are established, the algorithm simulates an outage contingency and re-runs the load flow analysis under these conditions. The power flows in all transmission lines and the maximum power flow, are calculated, alongside the voltage magnitudes at all buses. Also, the model computes the performance indices, including $PI_P$ and $PI_V$, to assess system stability. The OPI is then evaluated to classify the severity of each contingency. Then the model ranks the contingencies based on the OPI.

### C. Random Forests

Random forests (RF) is an ensemble learning technique that improves the performance of classification by building multiple decision trees during training and aggregating the outputs by majority decision [31]. The RF algorithm extends the traditional decision tree model by reducing overfitting and improving generalisation by bootstrap aggregation (bagging) and random selection of the features. RF is widely used in the contingency classification of energy systems and has proven to be reliable in handling high dimensional data and complex interactions of features [25].

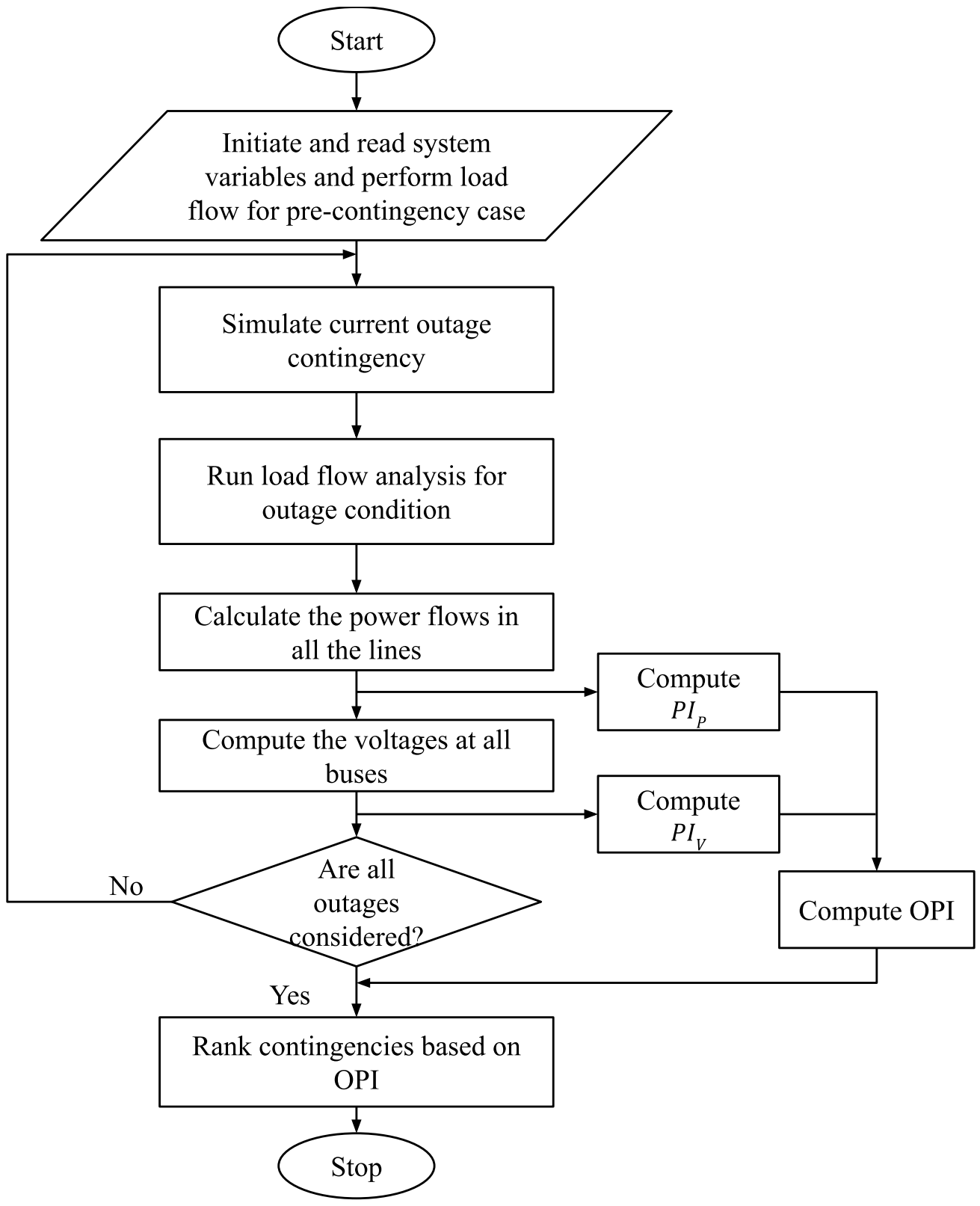

Figure 1: Flowchart of load flow algorithm

Mathematically, RF operates by constructing a set of $B$ decision trees, each trained on a different bootstrap sample of the original dataset. Given an input feature vector $X$, the prediction of an individual tree, denoted as $h_b(X)$, contributes to the final classification decision. The ensemble prediction is determined by majority voting across all trees expressed in (12).

$$Y^{\wedge} = mode\{h_1(X), h_2(X), ..., h_B(X)\} \quad (12)$$

where $h_b(X)$ represents the prediction of the $b^{th}$ tree and $Y^{\wedge}$ is the final predicted class.

The strength of RF lies in reducing variance, since the addition of more trees reduces the risk of overfitting. In addition, RF uses a random selection mechanism for the selection of features, with each tree being trained on a subset of features, which ensures the diversity and robustness of the model. The RF was used to prioritise the contingency scenarios, identify high risk failures and assess the stability of the network under different operational conditions [23]. Studies show that RF is excellent at capturing nonlinear relationships in contingency data sets, which makes it suitable for dynamic power systems environments.

### D. *Support Vector Machines*

Support Vector Machines (SVM) are supervised learning algorithms that seek to find the optimal hyperplane that maximizes the margin between different classes [32]. The use of SVM for power system contingency classification has been extensively investigated because of its ability to solve both linear and non-linear classification problems using kernel functions [26].

Mathematically, SVM aims to find a hyperplane of the form as expressed in (13)

$$w^T X + b = 0 \quad (13)$$

where $w$ is the weight vector, $X$ is the feature vector, and $b$ is the bias term. The optimal hyperplane is determined by maximizing the margin between the nearest data points (support vectors) of each class. The margin is defined in (14) which is maximized subject to the constraint expressed in (15).

$$\text{Margin} = \frac{2}{|w|} \quad (14)$$

$$y_i \left(w^T X_i + b\right) \geq 1, \quad \forall i \quad (15)$$

where $y_i$ represents the class label (either +1 or -1). For nonlinearly separable data, SVM utilizes kernel functions to transform data into a higher-dimensional space where a linear separation is possible. The commonly used Radial Basis Function (RBF) kernel is given by (16).

$$K(X_i, X_j) = e^{-\gamma |X_i - X_j|^2} \quad (16)$$

where $\gamma$ is a hyperparameter controlling the influence of individual training samples.

SVM is effectively used for detecting high risk failure events, assessing voltage stability and classifying contingency scenarios according to operational conditions. Studies have shown that SVM achieves high accuracy in classifying, especially when combined with selection techniques such as PCA [27].

### E. *K-Nearest Neighbors*

Unlike parametric algorithms, KNN is non-parametric and refrains from assuming any underlying data distribution. Instead, it depends on the data itself to generate predictions. Given a specific value of $x_t$, the KNN algorithm predicts the class as (17).

$$y_t = \arg \max_{x_i \in N(x_t, k)} E(y_t, x_i) \quad (17)$$

where $x_i$ is the new input to be tested and $y_t$ is the predicted class for the given new input, and $m$ is the number of classes presented in the training data. The function $E(a, b)$ is defined as (18).

$$E(a, b) = \begin{cases} 1, & \text{if } a = b \\ 0, & \text{otherwise} \end{cases} \quad (18)$$

and $N(x, k)$ represents the set of $k$ nearest neighbors of $x$. The KNN algorithm bases its class determination for the query instance solely on prior probabilities, without considering the class distribution in the vicinity of the query point.

## III. DATA PREPROCESSING AND MODELLING

This section describes the workflow around simulation and modelling to generate the dataset used in this research. In this study, the dataset was prepared and structured as a multi-class classification problem, with the contingencies classified into safe, moderate and severe. Given the imbalance in the distribution of contingencies (where major contingencies account for a much smaller proportion) different data pre-processing techniques have been used to improve the performance of the models and to ensure fair training.

### A. *Data Normalization*

In this work, to enhance the consistency of input features and improve model performance, normalization was applied to both the input and output parameters, denoted as $X$. The normalized value, $X_n$, was computed using the transformation in (19):

$$X_n = 0.8 \left( \frac{X - X_{min}}{X_{max} - X_{min}} \right) + 0.1 \quad (19)$$

where $X_{min}$ and $X_{max}$ are the minimum and maximum values of the original dataset. Normalization scales data to a range of 0.1 to 0.9 and thus preserves relative relationships between elements. Normalisation is of particular importance for distance algorithms such as KNN, since it ensures that all parameters contribute to the similarity measurement in a proportional way. In addition, it also benefits other machine learning algorithms by increasing the performance of numerical stability and convergence.

*B. Synthetic Minority Over-sampling Technique*

Due to the severe class imbalance in contingency classification, where the Severe contingency class is significantly underrepresented compared to the Safe and Moderate classes, the Synthetic Minority Over-sampling Technique (SMOTE) was applied. SMOTE operates by selecting k-nearest neighbors for a given minority class instance and generating synthetic points along the line segments joining the instance and its neighbors [34][27]. Mathematically, if $x_i$ is a randomly chosen instance from the minority class and $x_{zi}$ is one of its $k$ nearest neighbors, the synthetic sample $x_{new}$ is generated by (20).

$$x_{new} = x_i + \lambda(x_{zi} - x_i) \tag{20}$$

where $\lambda$ is a random number in the range [0, 1], ensuring that the new instance is placed somewhere along the vector connecting $x_i$ and $x_{zi}$ [23][28].

*C. Principal Component Analysis*

In reducing dimensionality and removing redundant or correlated features, Principal Component Analysis (PCA) was applied. PCA is a widely used technique in machine learning that transforms a dataset into a lower-dimensional space, preserving the most significant variance while improving computational efficiency [24].

PCA involves finding new orthogonal axes, called principal components, that maximize the variance in the dataset. PCA transformation follows these key steps: the data is first standardized to ensure that all features contribute equally, the covariance matrix of the dataset is calculated to capture relationships between features, eigenvalues and eigenvectors of the covariance matrix are computed. The eigenvectors represent the principal components, while the corresponding eigenvalues indicate importance. Finally, dimensionality reduction is performed; the top *k* principal components with the highest eigenvalues are selected, and the original data is projected onto this lower-dimensional subspace. Mathematically, given a dataset $X$ of $n \times m$ dimensions, PCA decomposes $X$ as expressed in (21).

$$X' = XW \tag{21}$$

where $W$ is a matrix of the top $k$ eigenvectors and $X'$ is the transformed dataset in the new lower-dimensional space.

*D. Data Preparation for Contingency Classification*

Efficient data preparation is critical for ensuring the accuracy and reliability of contingency classification in power systems. The flowchart in Fig. 1 provides a structured visualization of the key steps involved in preparing data for KNN, RF, and SVM models. Fig. 2 presents a block diagram outlining the data preparation pipeline for the ML models, detailing the key steps which include: data collection, class labeling based on normalized OPI values, and preprocessing techniques.

*1) Data Collection:* The methodology for generating data sets uses a systematic approach for simulating and analysing power system contingencies. The contingency scenarios include N-k contingency (where k ranges from 1 to 3) for IEEE-14 and IEEE-30 bus test systems, which provide a comprehensive framework for the assessment of the safety of power systems. The types of contingencies considered include power failure of transmission lines, generator failure and variations in load, occurring alone and in combination with others, to reflect real world disturbances of the system. For each test system, the methodology uses both exhaustive enumeration and intelligent sampling strategies for the generation of contingency scenarios. The simulation process starts with a base-case power flow solution and continues with the application of contingencies in order. The extracted features comprise a comprehensive of all power system parameters, including bus voltage magnitudes and angles, line loading percentages, active and reactive power flows, and generator output parameters. OPI is then calculated as a measure of the security of the system.

*2) Class Labeling:* To systematically categorize power system contingencies based on security impact, a three-class classification framework is adopted. The classification is determined based on the normalized OPI. The contingency classes are defined as follows: Safe-contingencies with normalized OPI value less than or equal to 0.2, Moderate-contingencies with a normalized OPI value between 0.2 and 0.5, and Severe-contingencies with a normalized OPI value greater than 0.5), indicating significant disturbances that could compromise system security and require immediate intervention. The normalized OPI variable is removed from the dataset to avoid data leakage, ensuring that models do not rely on predefined security metrics but rather learn from relevant power system parameters.

*3) Data preprocessing Techniques:* The dataset is examined for incomplete or missing data points. Missing values are addressed using sample removal. This step ensures data consistency and prevents biased model training. Additionally, input variables are normalized to a common range. This prevents any single feature from disproportionately influencing model predictions. To facilitate effective model learning and performance evaluation, the dataset is divided into training (80%) and validation (20%) sets. This split ratio ensures that the models generalize well to unseen contingencies while maintaining robust training.

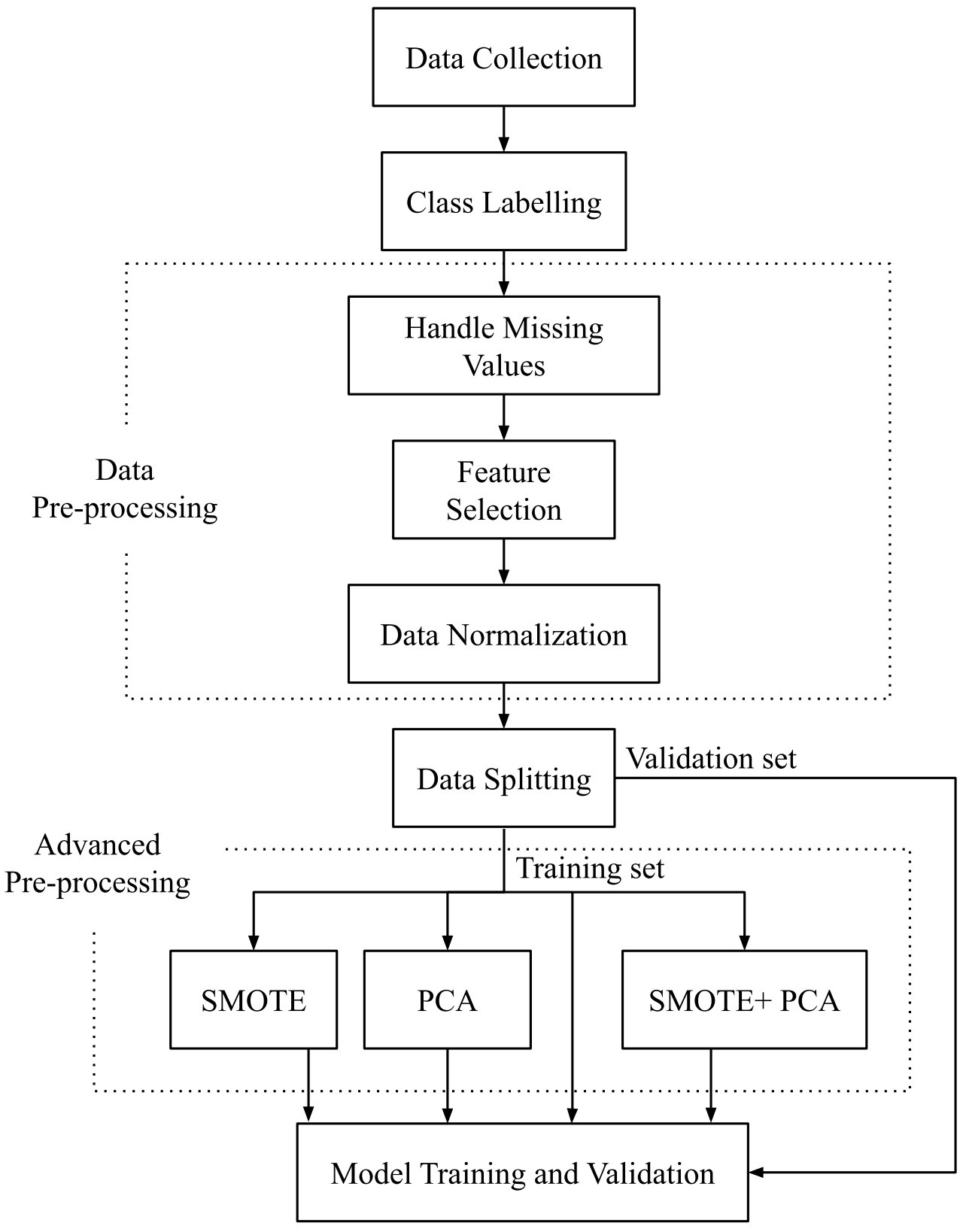

Fig. 2. Block diagram of data preparation processes

*4) Advanced Data Processing Techniques:* To enhance model performance, two advanced data preprocessing techniques—SMOTE and PCA—are applied on the training dataset. These methods address data imbalance and feature dimensionality reduction, respectively. The dataset is prepared in four distinct configurations to evaluate the impact of different preprocessing techniques:

a) *Normalized:* Dataset with only feature scaling applied.
b) *SMOTE-balanced:* Dataset with SMOTE applied to handle class imbalance.
c) *PCA-transformed:* Dataset where dimensionality reduction is performed.
d) *SMOTE + PCA-transformed:* Dataset where class balancing and feature reduction are jointly applied.

Each model is trained and evaluated on these four dataset configurations to assess the impact of preprocessing on classification accuracy.

### *E. Model Training and Evaluation*

The training and evaluation of the KNN, RF, and SVM models is conducted using four distinct variations of the dataset to assess the impact of different preprocessing strategies. These variations include a normalized dataset, where only feature scaling was applied; a SMOTE-balanced dataset, in which class imbalance was addressed using SMOTE; a PCA-transformed dataset, where dimensionality reduction was performed using PCA; and a SMOTE + PCA dataset, which incorporated both class balancing and dimensionality reduction. In this study, each model hyperparameters are selected through a random search approach to optimize performance, ensuring the best model configurations are chosen, trained, and tested across the four configurations to evaluate the effect of data preprocessing on classification performance.

Given the class imbalance in the dataset, accuracy was not used as the primary evaluation metric due to the skewed distribution of contingency classes. A high accuracy value could be misleading if the model predominantly classifies instances into the majority classes (Safe and Moderate) while failing to correctly identify Severe contingencies. Instead, the evaluation focused on Precision, Recall, and F1-score, particularly for the Severe contingency class. Precision measures the proportion of correctly identified Severe contingency cases out of all instances predicted as Severe and is mathematically expressed as (22).

$$P = \frac{TP}{TP + FP} \tag{22}$$

where $TP$ represents the number of correctly classified Severe contingencies and $FP$ denotes the number of incorrectly classified Severe cases. Recall quantifies the proportion of actual Severe contingencies that were correctly identified and is defined as (23)

$$R = \frac{TP}{TP + FN} \tag{23}$$

where $FN$ represents the number of Severe contingencies that were misclassified into other categories. Since both Precision and Recall alone may not fully capture the trade-offs between false positives and false negatives, the F1-score was utilized as a more balanced metric and is expressed in (24).

$$F1 = \frac{2PR}{P + R} \tag{24}$$

## IV. RESULTS AND FINDINGS

This section presents a comprehensive analysis of the performance of machine learning models—KNN, RF, and SVM—in classifying power system contingencies into three security levels: Safe, Moderate, and Severe. The evaluation aims to assess the impact of different data preprocessing techniques, including SMOTE and PCA on overall model performance.

### *A. Performance Comparison Across Models and Pre-processing Techniques*

To analyze the effect of different preprocessing techniques, four dataset configurations were utilized for model training before evaluation on testset: normalized, SMOTE-balanced, PCA-transformed, SMOTE + PCA transformed datasets. Each model is trained and tested on these variations to assess the impact of preprocessing on classification performance. Table I summarizes the distribution of contingency classes across dataset variations, as well as the dimensionality reduction effect of PCA.

### *B. Model performance on IEEE-14 Bus Dataset*

Table II presents the Precision, Recall, and F1-score for the severe class as evaluated across each classification model under the four IEEE-14 dataset configurations.

The RF model demonstrated high classification performance when trained on the Normalized Data and PCA-transformed Data, achieving F1-score of 0.86 in both cases. However, application of SMOTE significantly reduced the F1-score to 0.57, indicating that the synthetic data introduced by SMOTE negatively impacted the model's ability to distinguish Severe contingencies accurately. The worst performance was observed in the SMOTE + PCA dataset, with F1-score of 0.33, suggesting that the combination of class balancing and feature reduction introduced excessive variance, leading to poor classification reliability.

TABLE I. DISTRIBUTION OF CONTINGENCY CLASSES ACROSS DATASET CONFIGURATIONS

| Dataset | Configuration | Safe | Moderate | Severe | Feature Count |
|---|---|---|---|---|---|
| IEEE-14 Bus | Normalized | 3051 | 71 | 14 | 81 |
| | SMOTE-balanced | 3051 | 1000 | 1000 | 81 |
| | PCA-transformed | 3051 | 71 | 14 | 15 |
| | SMOTE + PCA | 3051 | 1000 | 1000 | 15 |
| IEEE-30 Bus | Normalized | 21316 | 2682 | 262 | 172 |
| | SMOTE-balanced | 21316 | 5000 | 1000 | 172 |
| | PCA-transformed | 21316 | 2682 | 262 | 20 |
| | SMOTE + PCA | 21316 | 5000 | 1000 | 20 |

TABLE II. PERFORMANCE OF MACHINE LEARNING MODELS ON IEEE-14 BUS DATASET

| Model | Dataset Configuration | Precision (Severe) | Recall (Severe) | F1-Score (Severe) |
|---|---|---|---|---|
| RF | Normalized | 1 | 0.75 | 0.86 |
| | SMOTE-balanced | 0.67 | 0.5 | 0.57 |
| | PCA-transformed | 1 | 0.75 | 0.86 |
| | SMOTE + PCA | 0.5 | 0.25 | 0.33 |
| SVM | Normalized | 0 | 0 | 0 |
| | SMOTE-balanced | 0 | 0 | 0 |
| | PCA-transformed | 1 | 0.5 | 0.67 |
| | SMOTE + PCA | 1 | 0.5 | 0.67 |
| KNN | Normalized | 0.75 | 0.75 | 0.75 |
| | SMOTE-balanced | 0.67 | 0.5 | 0.57 |
| | PCA-transformed | 0.67 | 0.5 | 0.57 |
| | SMOTE + PCA | 0.67 | 0.5 | 0.57 |

The SVM model failed to classify any Severe contingencies correctly when trained on both Normalized Data and SMOTE-balanced Data. However, when PCA was applied, SVM performance improved significantly, achieving an F1-score of 0.67. The same performance was observed for SMOTE + PCA, suggesting that while SVM benefits from feature reduction, the introduction of synthetic samples through SMOTE does not provide additional advantages.

The KNN model exhibited relatively stable performance across all dataset configurations, with F1-scores between 0.57 and 0.75. The best performance was achieved with Normalized Data, reinforcing the importance of feature scaling in distance-based models. The application of SMOTE and PCA did not lead to significant improvements, with F1-scores remaining constant across these variations. This suggests that KNN is less affected by class balancing or feature reduction compared to RF and SVM, but it still performs best when trained on raw normalized data without modifications.

TABLE III. PERFORMANCE OF MACHINE LEARNING MODELS ON IEEE-30 BUS DATASET

| Model | Dataset Configuration | Precision (Severe) | Recall (Severe) | F1-Score (Severe) |
|---|---|---|---|---|
| RF | Normalized | 0.98 | 0.95 | 0.97 |
| | SMOTE-balanced | 0.98 | 0.92 | 0.95 |
| | PCA-transformed | 0.97 | 0.97 | 0.97 |
| | SMOTE + PCA | 0.95 | 0.94 | 0.95 |
| SVM | Normalized | 0.98 | 0.82 | 0.89 |
| | SMOTE-balanced | 0.96 | 0.85 | 0.9 |
| | PCA-transformed | 0.88 | 0.94 | 0.91 |
| | SMOTE + PCA | 0.88 | 0.94 | 0.91 |
| KNN | Normalized | 0.98 | 0.69 | 0.81 |
| | SMOTE-balanced | 0.96 | 0.72 | 0.82 |
| | PCA-transformed | 0.92 | 0.75 | 0.83 |
| | SMOTE + PCA | 0.9 | 0.8 | 0.85 |

### *C. Model Performance on IEEE-30 Bus Dataset*

Table III presents the Precision, Recall, and F1-score for the severe class as evaluated across each classification model under the four IEEE-30 dataset configurations.

The RF model demonstrated high classification performance across all dataset configurations. When trained on Normalized Data and PCA-transformed Data, RF achieved an F1-score of 0.97, indicating strong generalization and reliability in identifying Severe contingencies. Applying SMOTE slightly reduced the F1-score to 0.95, suggesting that while class balancing helped improve Recall, it may have introduced minor false positives, reducing Precision. The SMOTE + PCA dataset yielded an F1-score of 0.95, confirming that RF remains highly robust across different preprocessing techniques, with minimal performance variation.

The SVM algorithm exhibited stable performance across different dataset configurations. Training on Normalized Data resulted in an F1-score of 0.89, while applying SMOTE improved the F1-score. PCA transformation led to a further increase in performance. The SMOTE + PCA configuration produced identical results, suggesting that SVM benefits more from feature selection rather than class balancing.

The KNN algorithm showed relatively lower performance compared to RF and SVM, particularly in Recall for Severe contingencies. When trained on Normalized Data, the F1-score was 0.81, indicating that distance-based classification struggled with class imbalance. The application of SMOTE improved the F1-score. PCA transformation increased the F1-score to 0.83, suggesting that reducing feature dimensionality improved classification efficiency. The best performance was observed in the SMOTE + PCA dataset, indicating that combining class balancing and dimensionality reduction resulted in the most effective configuration for KNN.

### *D. Discussion of Models Efficiency*

The results from the IEEE-14 and IEEE-30 bus systems provide comprehensive insights into the effectiveness of different machine learning models and data preprocessing techniques in contingency classification. Across all scenarios, RF emerged as the most robust model, consistently achieving high Precision, Recall, and F1-score across different dataset configurations. This reinforces the adaptability of RF in handling imbalanced datasets and learning complex relationships between power system parameters. The model performed particularly well on PCA-transformed data, indicating that dimensionality reduction helped eliminate redundant features while preserving key information for classification. However, applying SMOTE to RF did not yield significant improvements and, in some cases, slightly reduced Precision, suggesting that RF is inherently capable of managing class imbalance without requiring additional synthetic data. The confusion matrix in Fig. 3(a) visualizes the performance of the RF model on the IEEE-14 bus evaluation dataset, classifying contingencies into Safe, Moderate, and Severe categories, with true labels on the vertical axis and predicted labels on the horizontal axis, showing correct classifications of 761 Safe, 17 Moderate, and 3 Severe contingencies, alongside minimal misclassifications, highlighting strong accuracy. Similarly, the confusion matrix in Fig. 3(b) depicts the performance on the IEEE-30 bus evaluation dataset, featuring a dominant diagonal indicating high correct prediction rates across all classes with minimal misclassifications, validating its effectiveness on a more complex dataset.

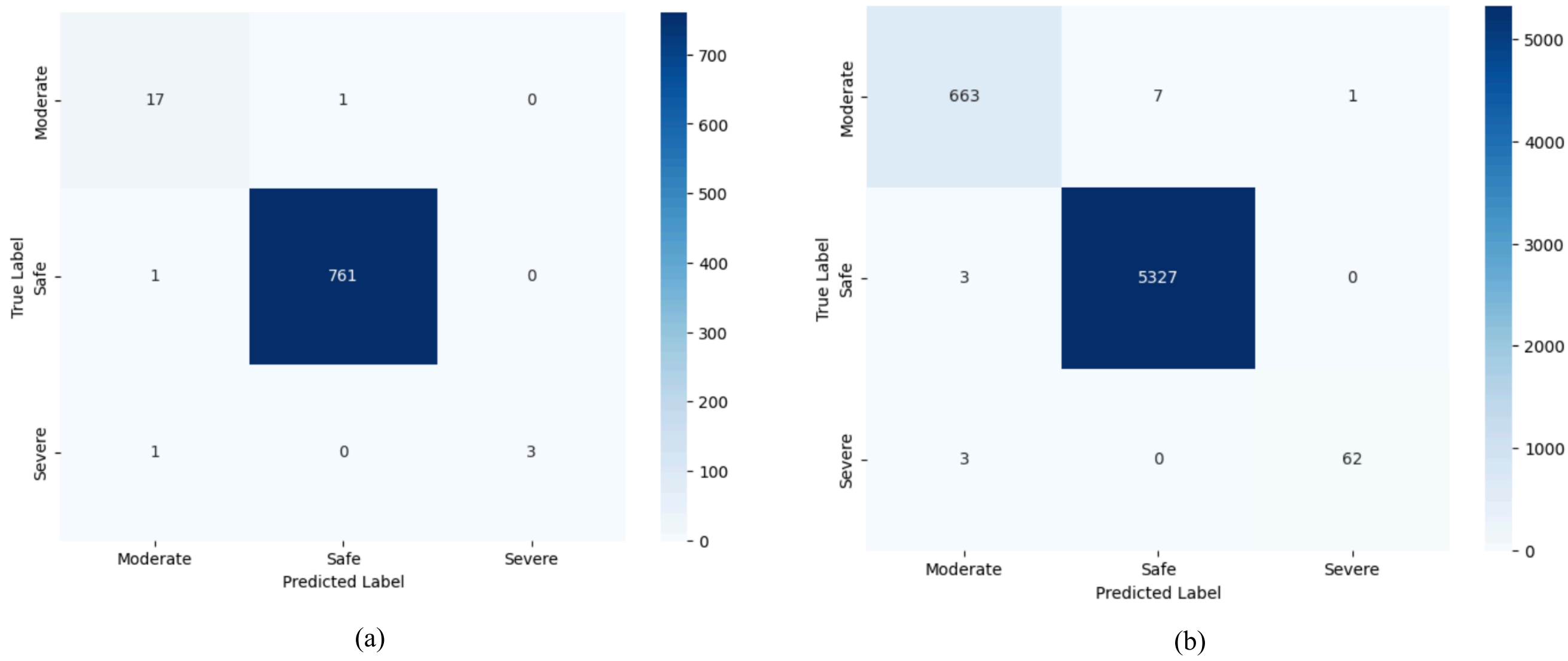

Fig. 3. Confusion matrix for random forest model on (a) IEEE-14 bus and (b) IEEE-30 bus evaluation dataset.

SVM exhibited mixed performance across different preprocessing approaches. When trained on raw normalized data, SVM struggled with class imbalance, particularly in the IEEE-14 bus system, where it failed to classify any Severe contingencies correctly due to its sensitivity to the severe class imbalance (only 14 Severe instances out of 3136 total instances) and high dimensionality (81 features), which hindered its ability to find an optimal hyperplane for separation. However, with PCA transformation reducing the feature count to 15, SVM performance significantly improved, achieving an F1-score of 0.67 for the Severe class, demonstrating that feature reduction was highly beneficial for this model. This improvement is likely due to SVM's reliance on optimal hyperplane separation, which benefits from a lower-dimensional space with minimized noise and redundancy. The application of SMOTE alone did not lead to major improvements in SVM performance, suggesting that feature selection played a more critical role than class balancing for this model. KNN exhibited the highest variability in performance, particularly struggling in the IEEE-14 bus system but demonstrating moderate improvements in the IEEE-30 bus system. As a distance-based model, KNN is highly sensitive to feature scaling and class distribution, which explains why normalized data led to the best results in the IEEE-14 bus system. In contrast, SMOTE and PCA in combination yielded the best performance in the IEEE-30 bus system, highlighting the importance of both class balancing and dimensionality reduction for KNN in large datasets. However, KNN generally underperformed compared to RF and SVM, suggesting that it may not be the most suitable model for high-dimensional contingency classification problems.

In assessing the impact of pre-processing techniques, it is clear that PCA has played a more important role than SMTP in improving the accuracy of the classification for the majority of models. RF and SVM have benefited significantly from the reduction of the number of elements, while KNN needed both SMOTE and PCA in order to achieve the best performance. SMOTE improved the recall of severe contingencies across models, but also resulted in a slight loss of precision, especially in RF, suggesting that the generation of synthetic data may introduce a certain amount of false positives. The combination of SMOTE and PCA has produced mixed results, with SMOTE showing a high benefit for KNN but offering little or no benefit for RF and SVM.

Overall, the IEEE-30 bus dataset demonstrated higher classification performance across models compared to the IEEE-14 bus system, most likely due to the larger sample space and comprehensive contingency scenarios available for training. The results confirm that RF remains the most reliable model for contingency classification, while SVM benefits significantly from feature selection, and KNN requires both class balancing and dimensionality reduction to perform optimally.

## CONCLUSION

This study shows the effectiveness of machine learning models for real-time classification of emergency situations in energy systems. Research is systematically assessing the impact of preprocessing techniques: SMOTE for class balancing and PCA for dimensionality reduction on the performance of IEEE-14 and IEEE-30 bus systems models.

The findings highlight that RF consistently outperforms other models and achieves high accuracy of classification across different data sets, especially when trained with transformed PCA data. SVM benefits significantly from reducing dimensionality, while KNN is best off with both class and function reductions. The results confirm that the choice of pre-processing techniques is of key importance for optimizing the performance of the model. While SMOTE improves the recall of underrepresented classes, it may introduce false positives, especially in the RF sector, which already has an effective way of dealing with class imbalances. However, the PCA has improved the computing power and the accuracy of the classification in RF and SVM. The combination of SMOTE and PCA has yielded mixed results, favouring KNN but not offering any significant benefits to RF or SVM. The integration of the Newton-Raphson load flow method with machine learning models provides a powerful framework for utilities and system operators to assess the safety of power systems rapidly and accurately in real time.

This study shows the adaptability of machine learning models to applications of real-time grid monitoring and risk assessment and the progress made in the methodology for energy system safety assessment. The reliance on simulated IEEE bus systems limits the practical significance, as these idealized models may not fully capture the complexities of real-world power systems, such as dynamic load variations or equipment degradation. Future research should prioritize validating the proposed machine learning models with real-world grid data to confirm their robustness and practical utility in operational environments. Additionally, expanding the training set to include diverse contingency scenarios in larger power systems, while exploring hybrid feature selection techniques, advanced methods like deep learning, and class balancing approaches such as ADASYN, will enhance the development of more robust, scalable, and adaptable systems for emergency assessment in modern power networks.